\documentclass[a4paper, 10pt, conference]{ieeeconf}      
\usepackage{FG2019}

\FGfinalcopy 

\IEEEoverridecommandlockouts                              
\usepackage{graphics} 
\usepackage{epsfig} 
\usepackage{lineno,hyperref}
\usepackage{amsmath} 
\usepackage{amssymb}  
\usepackage{algorithm}
\usepackage{caption}
\usepackage{subcaption}

\def\FGPaperID{****} 

\title{\LARGE \bf
Using Photorealistic Face Synthesis and Domain Adaptation to \\ Improve Facial Expression Analysis
}

\author{\parbox{16cm}{\centering
    {\large Behzad Bozorgtabar$^1$, Mohammad Saeed Rad$^1$, \\ 
    Haz{\i}m Kemal Ekenel$^{2}$, Jean-Philippe Thiran$^{1}$}\\
    {\normalsize
    $^1$\'Ecole Polytechnique F\'ed\'erale de Lausanne, Switzerland, $^{2}$Istanbul Technical University, Istanbul, Turkey}}
}

\begin{document}
\IEEEoverridecommandlockouts\pubid{\makebox[\columnwidth]{978-1-7281-0089-0/19/\$31.00~\copyright{}2019 IEEE \hfill}
\hspace{\columnsep}\makebox[\columnwidth]{ }}

\ifFGfinal
\thispagestyle{empty}
\pagestyle{empty}
\else
\author{Anonymous FG 2019 submission\\ Paper ID \FGPaperID \\}
\pagestyle{plain}
\fi
\maketitle

\begin{abstract}
Cross-domain synthesizing realistic faces to learn deep models has attracted increasing attention for facial expression analysis as it helps to improve the performance of expression recognition accuracy despite having small number of real training images. However, learning from synthetic face images can be problematic due to the distribution discrepancy between low-quality synthetic images and real face images and may not achieve the desired performance when the learned model applies to real world scenarios. To this end, we propose a new attribute guided face image synthesis to perform a translation between multiple image domains using a single model. In addition, we adopt the proposed model to learn from synthetic faces by matching the feature distributions between different domains while preserving each domain's characteristics. We evaluate the effectiveness of the proposed approach on several face datasets on generating realistic face images. We demonstrate that the expression recognition performance can be enhanced by benefiting from our face synthesis model. Moreover, we also conduct experiments on a near-infrared dataset containing facial expression videos of drivers to assess the performance using in-the-wild data for driver emotion recognition.
\end{abstract}

\section{INTRODUCTION}
\label{sec:intro}
Face image synthesis has received increasing attentions as it has practical applications in human-computer interactions, facial animation, face recognition, and facial expression recognition. The most notable solution in image synthesis was the incredible breakthroughs in generative models. In particular, Generative Adversarial Network (GAN) \cite{goodfellow2014generative} variants have achieved state-of-the-art results for image-to-image translation task. These GAN models could discover relations between two visual domain using paired \cite{isola2017image} or unpaired data \cite{kim2017learning,zhu2017unpaired} during training process. In addition, most existing GAN models \cite{shen2017learning,zhu2017unpaired} are proposed to synthesize images of a single attribute, which make their training inefficient in the case of having multiple attributes, since for each attribute a separate model is needed.

\begin{figure}
\centering
\begin{subfigure}[b]{0.45\textwidth}
   \includegraphics[width=1\linewidth]{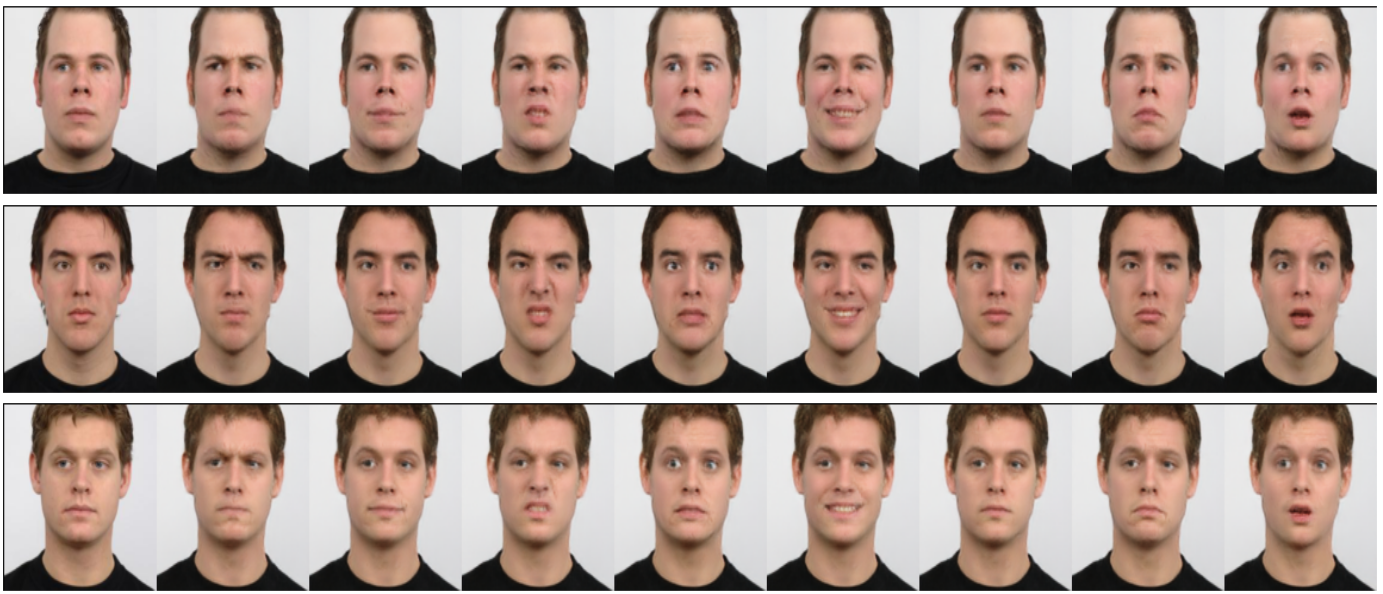}
   \caption{Photorealistic face synthesis}
   \label{fig1_1} 
\end{subfigure}

\begin{subfigure}[b]{0.45\textwidth}
   \includegraphics[width=1\linewidth]{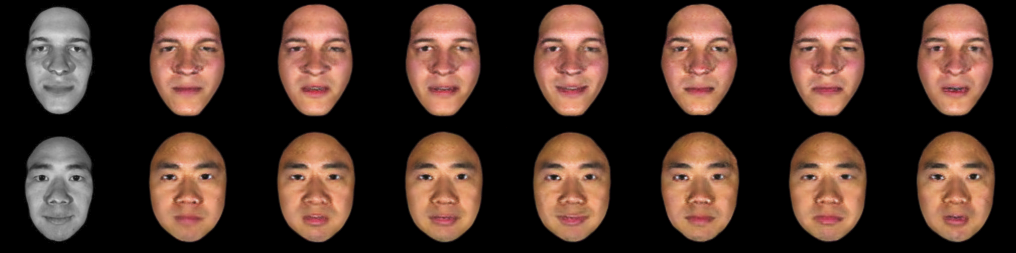}
   \caption{Realism refinement using domain adaptation}
   \label{fig1_2}
\end{subfigure}
\caption{Attribute guided face image generation using our method. (a) The input \textit{neutral} faces are fed into our model to exhibit specified attributes. \textbf{Left to right}: input \textit{neutral} face and eight face attributes including \textit{angry}, \textit{contemptuous}, \textit{disgusted}, \textit{fearful}, \textit{happiness}, \textit{neutral}, \textit{sadness} and \textit{surprised}, respectively. (b) Our model improves the realism of synthetic face (\textbf{left}), while preserving the identity and face pose information during the realism refinement for a specified attribute.}
\label{fig:1}
\end{figure}

In this paper, we pursue several objectives; synthesizing realistic faces by controlling the facial attributes of interest (e.g. face expression), preserving the facial identity after manipulation, and to investigate learning from synthetic facial images for improving expression recognition accuracy (see Fig. \ref{fig:1}). Our objective is to use a single model to synthesize face photos with a desired attribute and translate an input image from one domain into multiple domains without having matching image pairs. Our proposed method is based on encoder-decoder structure, using the image latent representation, where we model the shared latent representation across image domains. Therefore, during inference step, by changing face attributes, we can generate plausible face images owing attribute of interest. We also introduce bidirectional loss for the latent representation, which can resolve generator mode collapse to ensure diverse outputs can be produced depending on input attribute.
Our paper makes the following contributions: 
\begin{enumerate}
\item We extend the previous work \cite{BlindAuthors} and show that how the proposed approach can be used to generate photo-realistic facial images using synthetic face image and unlabeled real face images as the input. We compared our results with SimGAN method \cite{shrivastava2017learning} in terms of expression recognition accuracy to see improvement in the realism of generated faces using video data recorded in real world conditions;

\item Compared to other variants of GAN models \cite{zhu2017unpaired,perarnau2016invertible}, we demonstrate that the learnt representation achieves high-quality image synthesis results and preserves a certain expression that contribute to improve the performance of expression recognition accuracy to focus on the data augmentation process;

\item Lastly, unlike existing methods in face expression synthesis, which have only been validated on the face datasets captured in a lab-controlled environment, we tested our approach on the videos in the wild dataset, which contains arbitrary face poses, illumination and self-occlusions.
\end{enumerate}

\section{Related work}
\label{sec:related}
GAN based models \cite{goodfellow2014generative} have achieved impressive results in many image synthesis applications, including image super-resolution \cite{ledig2017photo}, image-to-image translation (pix2pix) \cite{isola2017image} and face image editing \cite{natsume2018rsgan, xiao2018elegant}. We summarize contributions of few important related works using GANs in the following subsections:

\subsection{Image-to-Image Translation}
Many of existing image-to-image translation methods e.g. \cite{isola2017image, shrivastava2017learning} formulated GANs in the supervised setting, where example image pairs are available. However, collecting paired training data can be difficult. On the other side, there are other GAN based methods, which do not require matching pairs of samples. For example, CycleGAN \cite{zhu2017unpaired} is capable to learn transformations from source to target domain without one-to-one mapping between two domain's training data. However, these GAN based methods could only train one specific model for each pair of image domains. Unlike the aforementioned approaches, we use a single model to synthesize multiple photo-realistic images, each having specific attribute. Recently, to manipulate attributes of image during image synthesis, conditional information, such as image labels, is used. As examples, IcGAN \cite{perarnau2016invertible} and StarGAN \cite{choi2018stargan} proposed image editing using AC-GAN \cite{odena2017conditional} with conditional information. However, we use domain adaptation by adding the realism to the simulated faces and there is no such a solution in these methods. Similar to \cite{perarnau2016invertible}, Fader Networks \cite{lample2017fader} proposed image synthesis model without needing to apply a GAN to the decoder output. However, these methods impose constraints on image latent space to enforce it to be independent from the attributes of interest, which may result in loss of information in generating attribute guided images.

\subsection{Face Attribute Manipulation and Face Generation}
Li et al. \cite{li2016deep} proposed a Deep convolutional network model for Identity-Aware Transfer (DIAT) of the facial attributes. The work \cite{shen2017learning} and \cite{kaneko2017generative} proposed to edit only single facial attribute. Lu et al. \cite{lu2018attribute} proposed attribute-guided face generation to translate low-resolution face images to high-resolution face images. Zhang et al. \cite{zhang2018generative} introduced the spatial attention mechanism into GAN framework, to only alter the attribute-specific face region and keep the rest unchanged.  Huang et al. \cite{huang2017beyond} proposed a Two-Pathway Generative Adversarial Network (TP-GAN) for photorealistic face synthesis by simultaneously considering local face details and global structures.

\subsection{Expression Transfer and Face Frontalization}
Zhang et al.\ \cite{zhang2018joint} proposed a method by disentangling the attributes (expression and pose) for simultaneous pose-invariant facial expression recognition and face images synthesis. Instead, we seek to learn attribute-invariant information in the latent space by imposing auxiliary classifier to classify the generated images. Zhou et al. \cite{zhou2017photorealistic} introduced a conditional difference adversarial autoencoder (CDAAE) to use emotion states as a conditional attribute for face generation. Lai et al. \cite{lai2018emotion} proposed a multi-task GAN-based network that learns to synthesize the frontal face images from profile face images. However, they require paired training data of frontal and profile faces. Instead, we seek to add realism to the synthetic frontal face images without requiring real frontal face images during training. Our method could generate realistic frontal face images using synthetic faces and real faces with arbitrary poses as input.

\section{Methods}
\label{sec:approach}
We first introduce our proposed attribute guided face synthesis model in Section \ref{subsec:attribute}. Then, we explore learning from simulated face images through adversarial training in Section \ref{subsec:posenormalization}. Finally, we discuss our implementation details and experimental results in Sections \ref{implementation} and \ref{subsec:experimentalresults}, respectively.

\subsection{Attribute-guided face image synthesis}
\label{subsec:attribute}
As the input of our framework, we have input face image, and the attributes to be edited (e.g. facial expression) and side image, which provides additional information to guide photo-realistic face synthesis. Let $\mathcal{X}$ and $\mathcal{S}$ denote original image and side conditional image domains, respectively and $\mathcal{Y}$ set of possible facial attributes. As the training set, we have $m$ triple inputs $\left (x_{i}\in \mathcal{X}, s_{i}\in \mathcal{S}, y_{i}\in \mathcal{Y} \right )$, where $x_{i}$ and $y_{i}$ are the $i^{th}$ input face image and binary attribute, respectively and $s_{i}$ represents the $i^{th}$ conditional side image. Then, for any categorical attribute vector $y$ from the set of possible facial attributes $\mathcal{Y}$, the objective is to train a model that will generate photo-realistic version (${x}'$ or ${s}'$) of the inputs ($x$ and $s$) from image domains $\mathcal{X}$ and $\mathcal{S}$ with desired attributes $y$.

\begin{figure*}
\centering
\includegraphics[height=9.5cm, width=18cm]{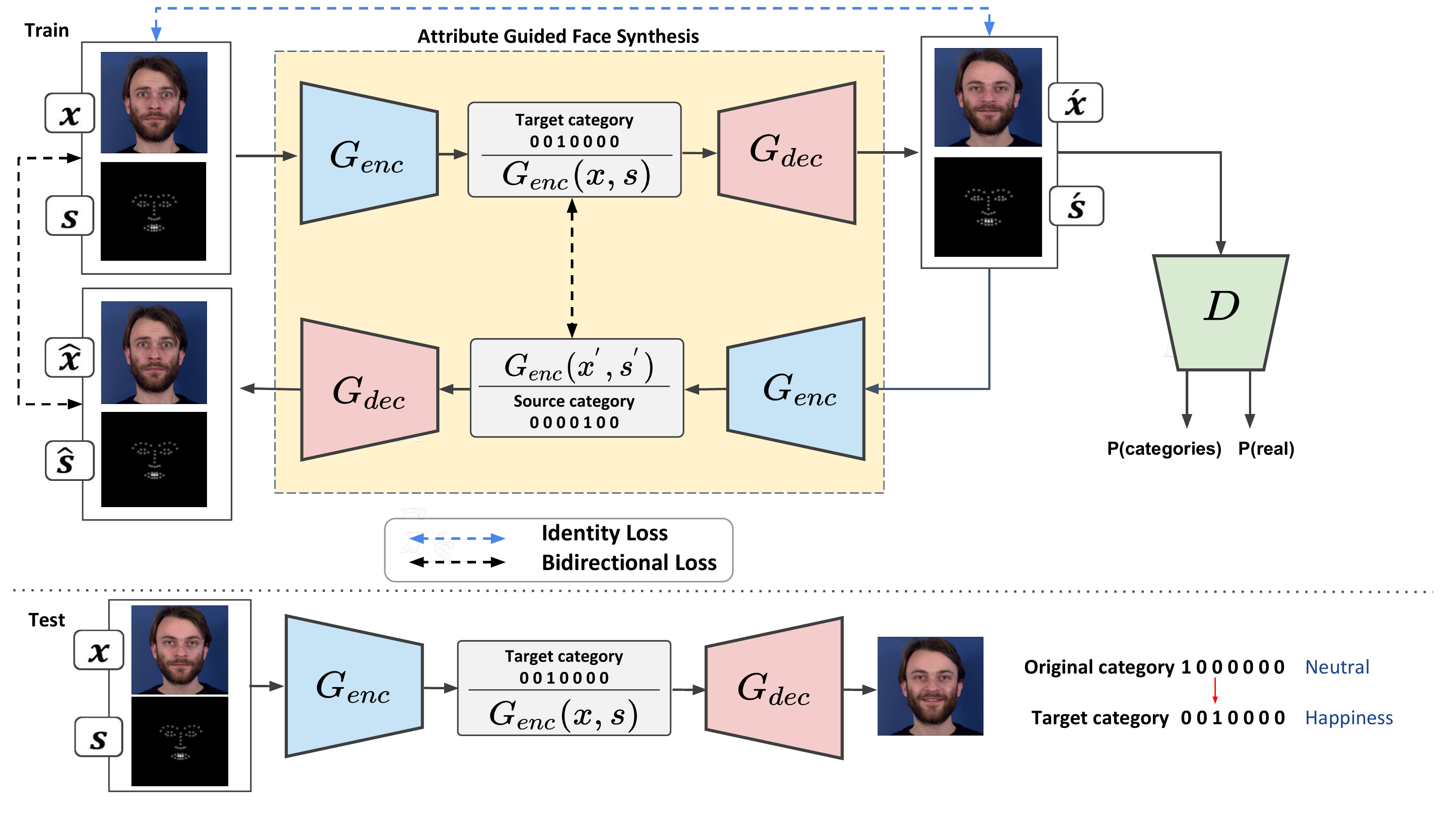}
  \caption{Overview of our proposed method. Training process of the attribute-guided face synthesis, consisting of two components, an encoder-decoder generator $G$ and a discriminator $D$ (\textbf{top}). Face expression synthesis at the test time (\textbf{bottom}).}
\label{fig:2}
\end{figure*}

Our model is based on the encoder-decoder architecture with domain adversarial training. As the input to our expression synthesis model (see Fig. \ref{fig:2}), we propose to incorporate individual-specific facial shape model (face landmark heatmap) as the side conditional information $s$ in addition to the original input image $x$. The face landmark heatmap contains 2D Gaussians centered at the landmarks locations, which are then simply concatenated channel-wise with the input image. Our goal is then to train a single generator $G$ with the encoder $G_{enc}$ -- decoder $G_{dec}$ networks to translate the input pair $\left ( x,s \right )$ from source domains into their corresponding output images $\left ( {x}',{s}' \right )$ in the target domain conditioned on the target domain attribute $y$ and the inputs latent representation $G_{enc}\left ( x,s \right )$, $G_{dec}\left ( G_{enc}\left ( x,s \right ),y \right )\rightarrow {x}',{s}'$. 
The encoder $G_{enc}:\left ( \mathcal{X}^{source}, \mathcal{S}^{source} \right )\rightarrow \mathbb{R}^{n\times \frac{h}{16}\times \frac{w}{16}}$ is a fully convolutional neural network with parameters $\theta _{enc}$ that encodes the input images into a low-dimensional feature space $G_{enc}\left ( x,s \right )$, where $n, h, w$ are the number of the feature channels and the input images dimensions, respectively. The decoder $G_{dec}:\left ( \mathbb{R}^{n\times \frac{h}{16}\times \frac{w}{16}},\mathcal{Y} \right )\rightarrow \mathcal{X}^{target}, \mathcal{S}^{target}$ is the sub-pixel \cite{shi2016real} convolutional neural network with parameters $\theta _{dec}$ that produce realistic images with target domain attribute $y$ and given the latent representation $G_{enc}\left ( x,s \right )$. The precise architectures of the neural networks are described in Section \ref{networkarchitechure}. During training, we randomly use a set of target domain attributes $y$ to make the generator more flexible in synthesizing images. In the following, we introduce the objectives for the proposed model optimization. 
\subsubsection{Adversarial Loss} 
We propose to learn attribute-invariant information in the latent space representing the shared features of the images sampled for different attributes. It means if the original and target domains are semantically similar (e.g. facial images of different expressions), we expect the common features across domains to be captured by the same latent representation. Then, the decoder must use the target attribute to do image-to-image translation from the original domain to the target domain. However, this learning process is unsupervised as for each training image from the source domain, its counterpart image in the target domain with attribute $y$ is unknown. Therefore, we propose to train an additional neural network called the discriminator $D$ (with the parameters $\theta _{dis}$) using an adversarial formulation to not only distinguish between real and fake generated images, but also to classify the image to its corresponding attribute categories. We use Wasserstein GAN \cite{gulrajani2017improved} objective with a gradient penalty loss $\mathcal{L}_{gp}$ \cite{arjovsky2017wasserstein} formulated as below:

\begin{equation} \label{eq1}
\begin{split}
\mathcal{L} _{adv}&=\mathbb{E}_{x,s}\left [ D_{src}\left ( x,s  \right ) \right ]\\
&-\mathbb{E}_{x,s,y}\left [ D_{src}\left ( G_{dec}\left ( G_{enc}\left ( x,s \right ),y \right ) \right ) \right ] 
-\lambda_{gp} \thinspace \mathcal{L}_{gp}\left ( D_{src} \right ),
\end{split}
\end{equation}
The term $D_{src}\left ( \cdot  \right )$ denotes a probability distribution over image sources given by $D$. The hyper-parameter $\lambda_{gp}$ is used to balance the GAN objective with the gradient penalty. A generator (encoder-decoder networks) used in our model has to play two roles: learning the attribute invariant representation for the input images and is trained to maximally fool the discriminator in a \textit{min-max} game. On the other hand, the discriminator simultaneously seeks to identify the fake examples for each attribute.

\subsubsection{Attribute Classification Loss} We use a classifier by returning additional output from the discriminator to perform an auxiliary task of classifying the synthesized and the real facial images into their respective attribute values. An attribute classification loss of real images $\mathcal{L}_{cls_{r}}$ to optimize the discriminator parameters $\theta _{dis}$ is defined as follow:
\begin{equation} \label{eq2}
\begin{split}
\min\limits_{\theta _{dis}}\mathcal{L}_{cls_{r}}& =\mathbb{E}_{x,s,{y}'}\left [ \ell_{r}\left ( x,s,{y}' \right ) \right ],\\
\ell_{r}\left ( x,s,{y}' \right )& =\sum_{i=1}^{m}-{y_{i}}'\log D_{cls}\left ( x,s \right )\\
&-\left ( 1-{y_{i}}' \right )\log\left ( 1-D_{cls}\left ( x,s \right ) \right ),
\end{split}
\end{equation}
Here, ${y}'$ denotes original attributes categories for the real images. $\ell_{r}$ is is the summation of binary cross-entropy losses of all attributes. Besides, an attribute classification loss of fake images $\mathcal{L}_{cls_{f}}$ used to optimize the generator parameters $\left ( \theta _{enc},\theta _{dec} \right )$, formulated as follow:
\begin{equation} \label{eq3}
\begin{split}
\min\limits_{\theta _{enc},\theta _{dec}}\mathcal{L}_{cls_{f}}& =\mathbb{E}_{x,s,{y}'}\left [ \ell_{f}\left ( {x}',{s}',y \right ) \right ],\\
\ell_{f}\left ( {x}',{s}',y \right )& =\sum_{i=1}^{m}-y_{i}\log D_{cls}\left ( {x}',{s}' \right )\\
&-\left ( 1-y_{i} \right )\log\left ( 1-D_{cls}\left ( {x}',{s}' \right ) \right ),
\end{split}
\end{equation}
where ${x}'$ and ${s}'$ are the generated images and auxiliary outputs, which should correctly own the target domain attributes $y$. $\ell_{f}$ denotes summing up the cross-entropy losses of all fake images.

\subsubsection{Identity Loss} Using the identity loss, we aim to preserve the attribute-excluding facial image details such as facial identity before and after image translation. We use a pixel-wise $l_{1}$ loss to enforce the facial details consistency and suppress the blurriness:
\begin{equation} \label{eq4}
\begin{split}
\mathcal{L}_{id} =
\mathbb{E}_{x,s,{y}'}\left [\left \|  G_{dec}\left ( G_{enc}\left ( x,s \right ),{y}' \right )-x\right \|_{1}  \right ],
\end{split}
\end{equation}

\subsubsection{Bidirectional Loss} 
Using adversarial loss alone usually leads to mode collapse, and the generator learns to ignore the attributes and changing $y$ at the test time could not generate diverse facial images. This issue has been observed in various applications of conditional GANs \cite{isola2017image,dosovitskiy2016generating} and to our knowledge, there is still no proper approach to deal with this issue. To address this problem, we show that using the trained generator, images of different domains can be translated bidirectionally. We decompose this objective into two terms: a bidirectional loss for the image latent representation, and a bidirectional loss between synthesized images and original input images, respectively. This objective is formulated using $l_{1}$ loss as follow:
\begin{equation} \label{eq5}
\begin{split}
\mathcal{L} _{bi} =\mathbb{E}_{x,s,{y}'}\left [\left \| x-\hat{x} \right \|_{1}+\left \| s-\hat{s} \right \|_{1}  \right ]+\\
\mathbb{E}_{x,s,y}\left [\left \| G_{enc}\left ( x,s \right )-G_{enc}\left ( {x}',{s}' \right ) \right \|_{1}  \right ], \\
{x}',{s}'=G_{dec}\left ( G_{enc}\left ( x,s \right ),y \right ), \\
\hat{x},\hat{s} =G_{dec}\left ( G_{enc}\left ( {x}',{s}' \right ),{y}' \right ),
\end{split}
\end{equation}

In the above equation, $\hat{x}$ and $\hat{s}$ denote the reconstructed original image and the side conditional image, respectively. Unlike \cite{zhu2017unpaired}, where only the cycle consistency losses are used at the image level, we additionally seek to minimize the reconstruction loss using latent representation.

\subsubsection{Overall Objective} Finally, the generator $G$ is trained with a linear combination of four loss terms: adversarial loss, attribute classification loss for the fake images, bidirectional loss, and identity loss. Meanwhile, the discriminator $D$ is optimized using an adversarial loss and attribute classification loss for the real images:

\begin{equation} \label{eq6}
\begin{split}
\mathcal{L}_{G}&=\mathcal{L}_{adv}+\lambda _{bi}\mathcal{L}_{bi}+\lambda _{cls}\mathcal{L}_{cls_{f}}+\lambda _{id}\mathcal{L}_{id},\\
\mathcal{L}_{D}&=-\mathcal{L}_{adv}+\lambda _{cls}\mathcal{L}_{cls_{r}},
\end{split}
\end{equation}
where $\lambda _{bi}$, $\lambda _{id}$ and $\lambda _{cls}$ are hyper-parameters, which tune the importance of identity loss, bidirectional loss and attribute classification loss, respectively.

\subsection{Realism Refinement Using Domain Adaptation}
\label{subsec:posenormalization}
\begin{figure}[h]
\centering
\includegraphics[height=5.5cm, width=9.3cm]{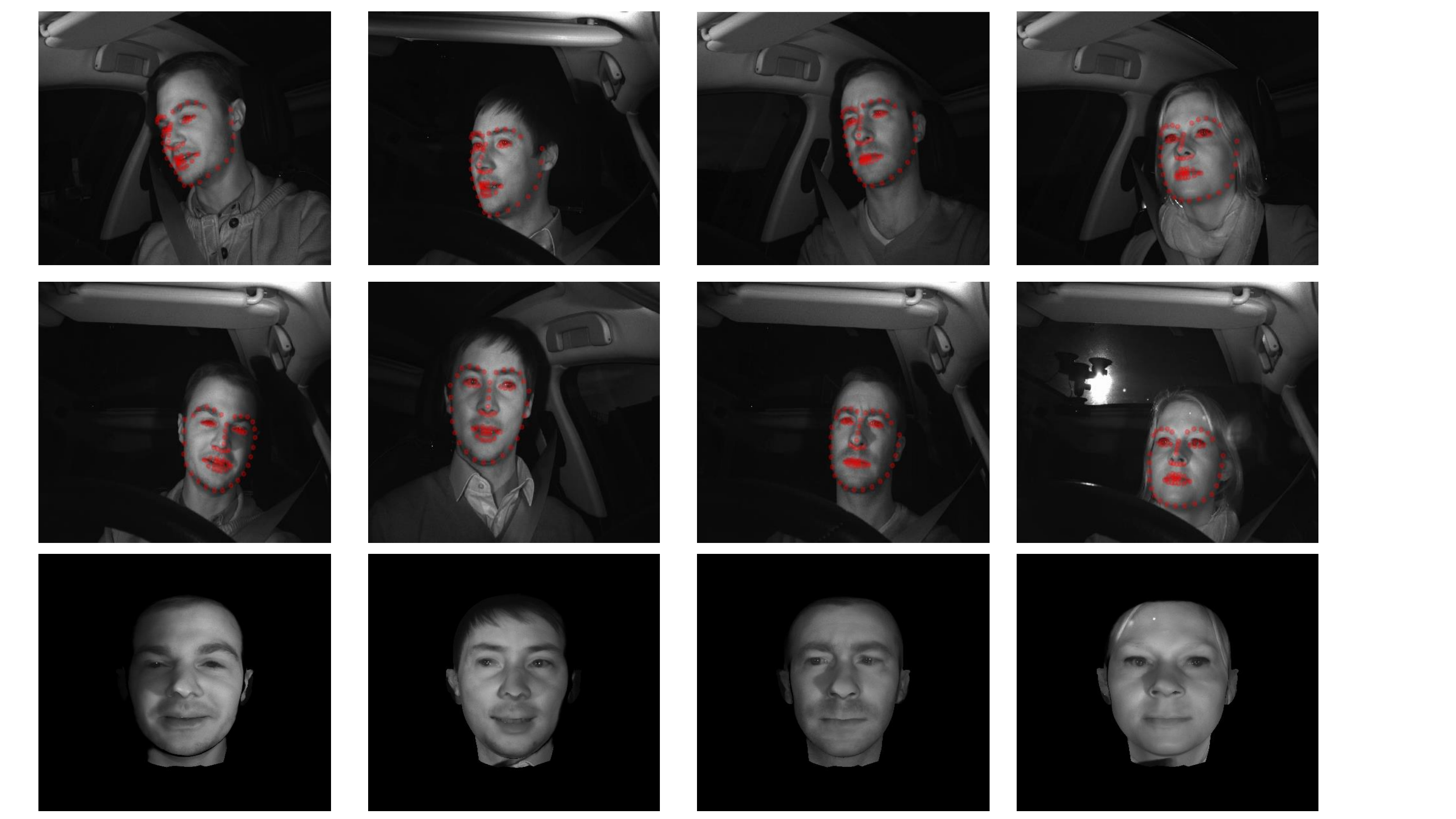}
  \caption{Examples of face frontalization process \cite{vlasic2005face} using facial landmarks. A simulated frontal view (\textbf{bottom row}) is generated from two side cameras (\textbf{top and middle rows}).}
\label{fig:3}
\end{figure}

\begin{figure}[h]
\centering
\includegraphics[height=5.5cm, width=8.3cm]{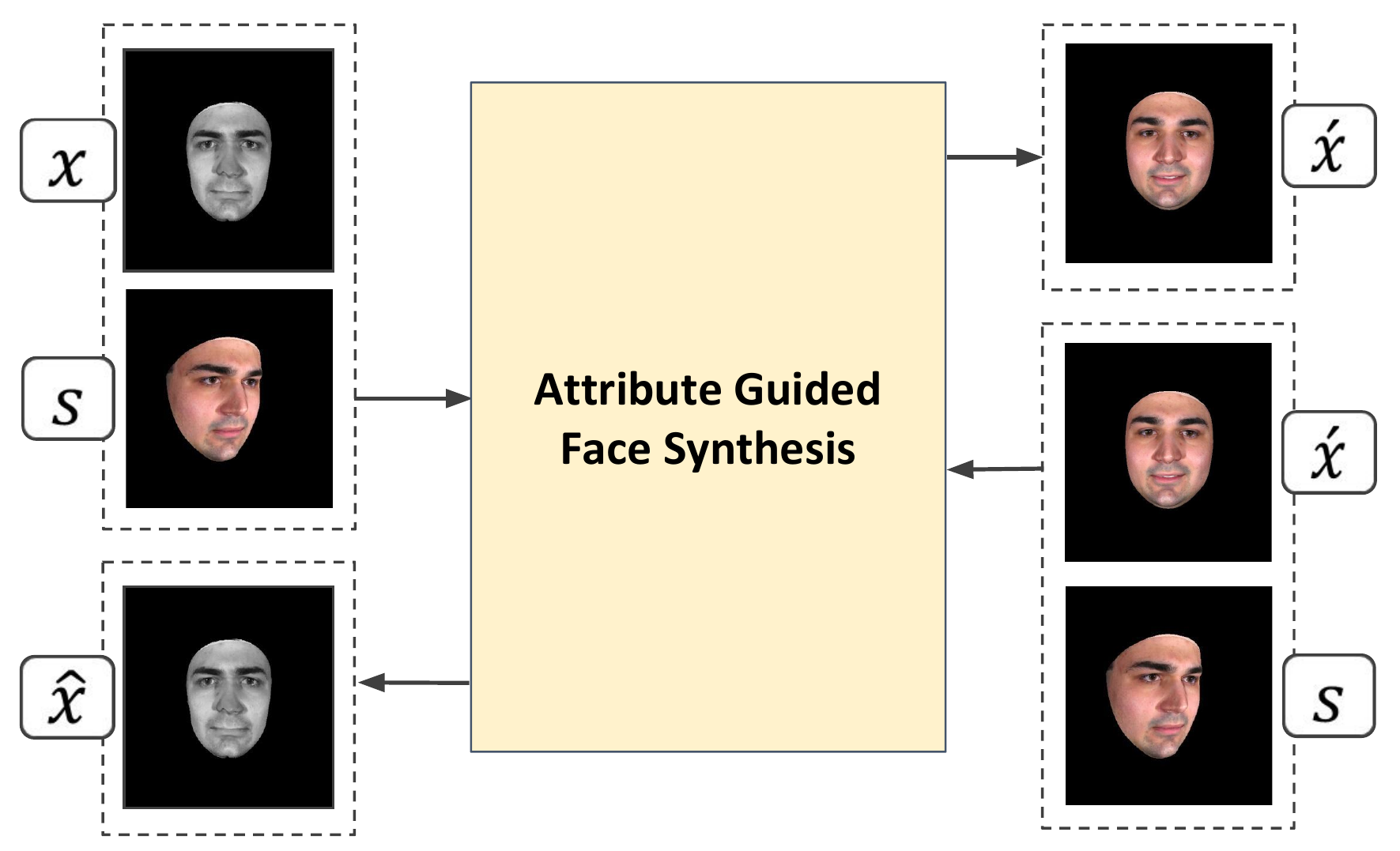}
  \caption{Overview of our proposed realism refinement. As the inputs, takes a simulated frontal face image and unlabeled real image as input and generates photo-realistic version of the frontal face.}
\label{fig:4}
\end{figure}

In an unconstrained face expression recognition (FER), accuracy will drop significantly for large pose variations. One possible solution would be using simulated faces rendered in frontal
view. In particular, we utilize a 3D Morphable Model using bilinear face model \cite{vlasic2005face} to construct a simulated frontal face image. Fig. \ref{fig:3} shows examples of simulated faces. However, learning from synthetic face images can be problematic due to a distribution discrepancy between real and synthetic images. Using proposed attribute guided face synthesis in Section \ref{subsec:attribute}, the model takes simulated frontal face image $x$ and real face image with arbitrary pose $s$ as inputs, and generates photo-realistic version of the synthetic face ${x}'$ during the realism refinement. With a side input image as condition, the model has enough information about the appearance of the desired face in advance and we transfer a texture from a given unlabeled real face image with arbitrary pose to a synthetic frontal face (see Fig. \ref{fig:4}). Here, the discriminator's role is to discriminate the realism of single generated image using real profile face images. In addition, using the same discriminator, we can generate images exhibiting arbitrary attributes e.g., different expressions.

We compare the pose-normalized face attribute transfer results of our proposed method with SimGAN method \cite{shrivastava2017learning} on the BU-3DFE dataset \cite{yin20063d}. SimGAN method \cite{shrivastava2017learning} considers learning from simulated and unsupervised images through adversarial training. Our method differs from SimGAN in following aspects: 1) we aim to synthesize photo-realistic frontal faces by preserving the face pose to address the challenges in unconstrained face expression recognition, whereas SimGAN is designed for simpler scenarios e.g., eye image refinement. 2) Another shortcoming of this method would be to ignore categorical information, which limits its performance. In contrast, our proposed method overcomes this issue by introducing attribute classification loss into our objective function. For a fair comparison with SimGAN method, we add the attribute classification loss by modifying the SimGAN's  discriminator, while keeping the rest of network unchanged. We achieve more visually pleasing results on test data compared to the SimGAN method (see Fig. \ref{fig:6}). Our proposed method can preserve the face image content while modifying only the attribute-related part of the images using the latent representation.
\section{Implementation Details}
\label{implementation}
All networks are trained using Adam optimizer \cite{kingma2014adam} $\left ( \beta _{1}=0.5,\beta _{2}=0.999 \right )$ and with a base learning rate of $0.0001$. We linearly decay learning rate after the first 100 epochs. We use a simple data augmentation with only flipping the images horizontally. The input image size and the batch size are set to $128\times 128$ and 8 for all experiments, respectively. We update the discriminator five times for each generator (encoder-decoder) update. The hyper-parameters in Eq. \ref{eq6} and Eq. \ref{eq1} are set as: $\lambda _{bi}=10$ and $\lambda _{id}=10$, $\lambda_{gp}=10$ and $\lambda _{cls}=1$, respectively. The whole model is implemented using PyTorch on a single NVIDIA GeForce GTX 1080.

\subsection{Network Architecture}
\label{networkarchitechure}
Tables \ref{table1:archidetails} and \ref{table2:archidetails} demonstrate the detailed network architectures of our proposed attribute-guided face image synthesis model. For the discriminator, we use PatchGAN \cite{isola2017image} that penalizes structure at the scale of image patches. Regarding the generator's decoder, we use sub-pixel convolution instead of transposed convolution followed by instance normalization \cite{ba2016layer}. Our experiments verify that it works remarkably better than transposed convolution for the face image synthesis.
\begin{table}[h]
\centering
\caption{The generator architecture. There are some notations; $n_{y}$ denotes the the dimension of domain attributes. IN and RB denote instance normalization and residual block, respectively.}
\label{table1:archidetails}
\resizebox{0.45\textwidth}{!}{ 
\begin{tabular}{ c c c c c c }
\hline\noalign{\smallskip}
Part & Layers & Input Size $\rightarrow $ Output Size &  Filter Size & Stride & Padding \\
\noalign{\smallskip}
\hline
\noalign{\smallskip}
  & {Conv+IN+ReLU} & $\left ( h,w,6 \right )\rightarrow \left ( h,w,64 \right )$ & $7\times7$ & 1 & 3  \\
  & {Conv+IN+ReLU} & $\left ( h,w,64 \right ) \rightarrow \left ( \frac{h}{2},\frac{w}{2},128 \right )$ & $4\times4$ & 2 & 1\\
\textbf{Encoder}  & {Conv+IN+ReLU} & $\left ( \frac{h}{2},\frac{w}{2},128 \right )\rightarrow \left ( \frac{h}{4},\frac{w}{4},256 \right )$ & $4\times4$ & 2 & 1  \\
  & {Conv+IN+ReLU} & $\left ( \frac{h}{4},\frac{w}{4},256 \right ) \rightarrow \left ( \frac{h}{8},\frac{w}{8},512 \right )$ & $4\times4$ & 2 & 1  \\
  & {Conv+IN+ReLU} & $\left ( \frac{h}{8},\frac{w}{8},512 \right ) \rightarrow \left ( \frac{h}{16},\frac{w}{16},1024 \right )$ & $4\times4$ & 2 &  1\\
  \hline
 & {RB:Conv+IN+ReLU} & $\left ( \frac{h}{16},\frac{w}{16},1024 \right ) \rightarrow \left ( \frac{h}{16},\frac{w}{16},1024 \right )$ & $3\times3$ & 1 & 1 \\
  & {RB:Conv+IN+ReLU} & $\left ( \frac{h}{16},\frac{w}{16},1024 \right ) \rightarrow \left ( \frac{h}{16},\frac{w}{16},1024 \right )$ & $3\times3$ & 1 & 1 \\
Encoder   & {RB:Conv+IN+ReLU} & $\left ( \frac{h}{16},\frac{w}{16},1024 \right ) \rightarrow \left ( \frac{h}{16},\frac{w}{16},1024 \right )$ & $3\times3$ & 1 & 1 \\
Bottleneck    & {RB:Conv+IN+ReLU} & $\left ( \frac{h}{16},\frac{w}{16},1024 \right ) \rightarrow \left ( \frac{h}{16},\frac{w}{16},1024 \right )$ & $3\times3$ & 1 & 1 \\
     & {RB:Conv+IN+ReLU} & $\left ( \frac{h}{16},\frac{w}{16},1024 \right ) \rightarrow \left ( \frac{h}{16},\frac{w}{16},1024 \right )$ & $3\times3$ & 1 & 1 \\
      & {RB:Conv+IN+ReLU} & $\left ( \frac{h}{16},\frac{w}{16},1024 \right ) \rightarrow \left ( \frac{h}{16},\frac{w}{16},1024 \right )$ & $3\times3$ & 1 & 1 \\
\noalign{\smallskip}
\hline
\hline
\noalign{\smallskip}
      & {Sub-Pixel Conv+IN+ReLU} & $\left ( \frac{h}{16},\frac{w}{16},1024+n_{y} \right ) \rightarrow \left ( \frac{h}{8},\frac{w}{8},512 \right )$ & $3\times3$ & 2 & 1 \\
            & {Sub-Pixel Conv+IN+ReLU} & $\left ( \frac{h}{8},\frac{w}{8},512 \right ) \rightarrow \left ( \frac{h}{4},\frac{w}{4},256 \right )$ & $3\times3$ & 2 & 1 \\
\textbf{ Decoder}                 & {Sub-Pixel Conv+IN+ReLU} & $\left ( \frac{h}{4},\frac{w}{4},256 \right ) \rightarrow \left ( \frac{h}{2},\frac{w}{2},128 \right )$ & $3\times3$ & 2 & 1 \\
                        & {Sub-Pixel Conv+IN+ReLU} & $\left ( \frac{h}{2},\frac{w}{2},128 \right ) \rightarrow \left ( h,w,64 \right )$ & $3\times3$ & 2 & 1 \\
& {\textbf{Image output}:Conv+Tanh} & $\left ( h,w,64 \right ) \rightarrow \left ( h,w,3 \right )$ & $7\times7$ & 1 & 3 \\
& {\textbf{Side output}:Conv+Tanh} & $\left ( h,w,64 \right ) \rightarrow \left ( h,w,3 \right )$ & $7\times7$ & 1 & 3 \\

\hline
\end{tabular}
}
\vspace*{0mm}
\end{table}

\begin{table}[h]
\centering
\caption{The discriminator architecture. FC and m denote fully connected layer and the number of target attributes, respectively}
\label{table2:archidetails}
\resizebox{0.45\textwidth}{!}{ 
\begin{tabular}{ c c c c c c }
\hline\noalign{\smallskip}
Part & Layers & Input Size $\rightarrow $ Output Size &  Filter Size & Stride & Padding \\
\noalign{\smallskip}
\hline
\noalign{\smallskip}
  & {Conv+Leaky ReLU} & $\left ( h,w,6 \right )\rightarrow \left ( \frac{h}{2},\frac{w}{2},64 \right )$ & $4\times4$ & 2 & 1  \\
  & {Conv+Leaky ReLU} & $\left ( \frac{h}{2},\frac{w}{2},64 \right ) \rightarrow \left ( \frac{h}{4},\frac{w}{4},128 \right )$ & $4\times4$ & 2 & 1\\
Discriminator  & {Conv+Leaky ReLU} & $\left ( \frac{h}{4},\frac{w}{4},128 \right )\rightarrow \left ( \frac{h}{8},\frac{w}{8},256 \right )$ & $4\times4$ & 2 & 1  \\
Hidden Layers  & {Conv+Leaky ReLU} & $\left ( \frac{h}{8},\frac{w}{8},256 \right ) \rightarrow \left ( \frac{h}{16},\frac{w}{16},512 \right )$ & $4\times4$ & 2 & 1  \\
  & {Conv+Leaky ReLU} & $\left ( \frac{h}{16},\frac{w}{16},512 \right ) \rightarrow \left ( \frac{h}{32},\frac{w}{32},1024 \right )$ & $4\times4$ & 2 &  1\\
 & {Conv+Leaky ReLU} & $\left ( \frac{h}{32},\frac{w}{32},1024 \right ) \rightarrow \left ( \frac{h}{64},\frac{w}{64},2048 \right )$ & $4\times4$ & 2 & 1 \\

\hline
\hline
\noalign{\smallskip}
Outputs& {\textbf{Output Layer}:Conv} & $\left ( \frac{h}{64},\frac{w}{64},2048 \right ) \rightarrow \left ( \frac{h}{64},\frac{w}{64},1 \right )$ & $3\times3$ & 1 & 1 \\
& {\textbf{Output Layer}:FC} & $\left ( \frac{h}{64},\frac{w}{64},2048 \right ) \rightarrow FCm$ & $-$ & $-$ & $-$ \\

\noalign{\smallskip}
\hline
\end{tabular}
}
\vspace*{0mm}
\end{table}
\section{Experimental Results}
\label{subsec:experimentalresults}
\subsection{Datasets}
\textbf{Near IR Drivers' Video Dataset}: We introduce the Near IR dataset that contains videos of emotion data captured from 26 subjects driving the cars in the multiple camera setup. This dataset is collected to support drivers by Advanced Driver Assistance Systems (ADAS). The drivers show six basic facial expressions including anger, disgust, fear, happiness, sadness, surprise plus neutral faces. In our experiments, we use frames (peak expressions) of 20 subjects for training and validation, and 6 subjects for the test, respectively.

\textbf{Oulu-CASIA VIS \cite{zhao2011facial}}: This dataset contains 480 sequences (from 80 subjects) of six basic facial expressions under the visible (VIS) normal illumination conditions. We conducted our experiments using subject-independent 10-fold cross-validation strategy.

\textbf{MUG \cite{aifanti2010mug}}: The MUG dataset contains image sequences of seven different facial expressions belonging to 86 subjects comprising 51 men and 35 women. The image sequences were captured with a resolution of $896\times 896$. We used image sequences of 52 subjects and the corresponding annotation, which are available publicly via the internet.

\textbf{BU-3DFE \cite{yin20063d}}: The Binghamton University 3D Facial Expression Database (BU-3DFE) \cite{yin20063d} contains 3D models from 100 subjects, 56 females and 44 males. The subjects show a neutral face as well as six basic facial expressions and at four different intensity levels. Following the setting in \cite{tariq2013maximum} and \cite{zhang2018joint}, we used an openGL based tool from the database creators to render multiple views from 3D models in seven pan angles $\left ( 0^{\circ},\pm 15^{\circ},\pm 30^{\circ},\pm 45^{\circ} \right )$.

\textbf{RaFD \cite{langner2010presentation} }: The Radboud Faces Database (RaFD) contains 4,824 images belonging to 67 participants. Each subject makes eight facial expressions.

\subsubsection{Qualitative evaluation} 
\begin{figure}[h]
\centering
\includegraphics[height=5.2cm, width=8.5cm]{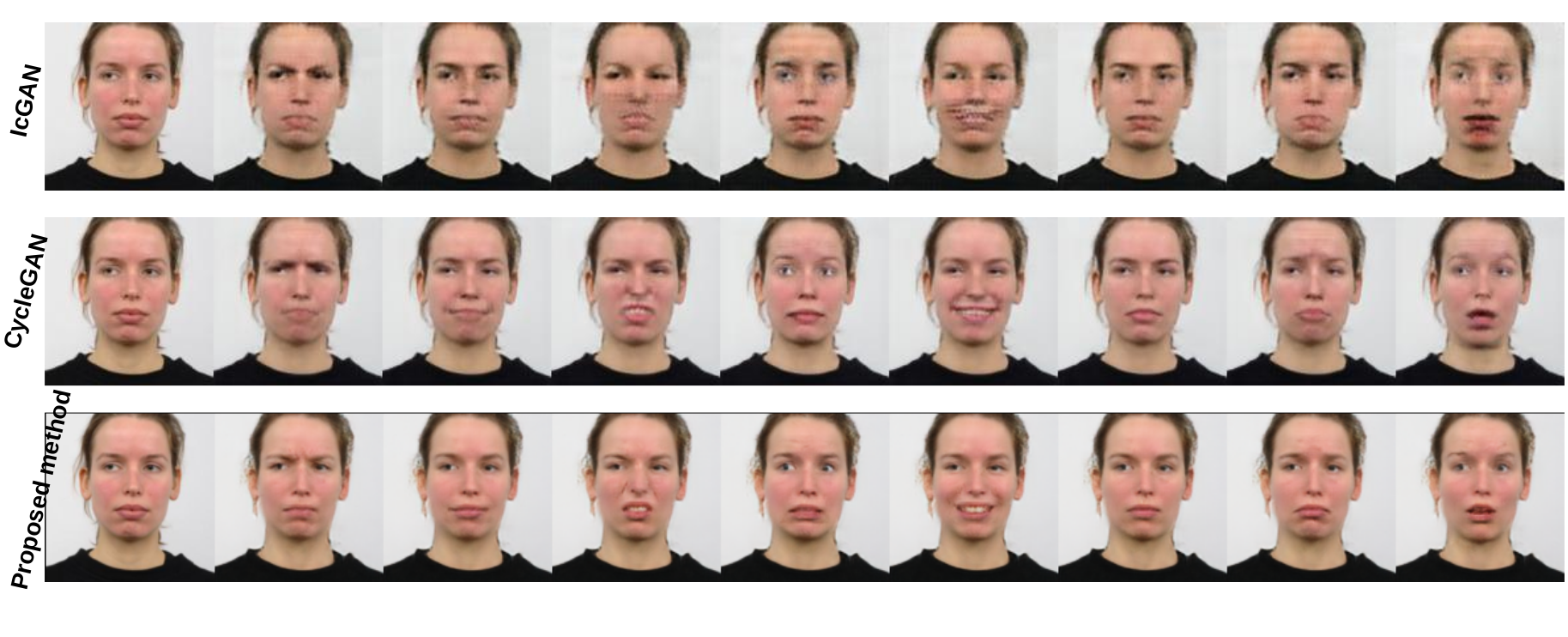}
  \caption{Facial attribute transfer results of our method compared with IcGAN \cite{perarnau2016invertible} and CycleGAN \cite{zhu2017unpaired}, respectively.}
\label{fig:5}
\end{figure}

\begin{figure}[h]
\centering
\includegraphics[height=6.5cm, width=8.2cm]{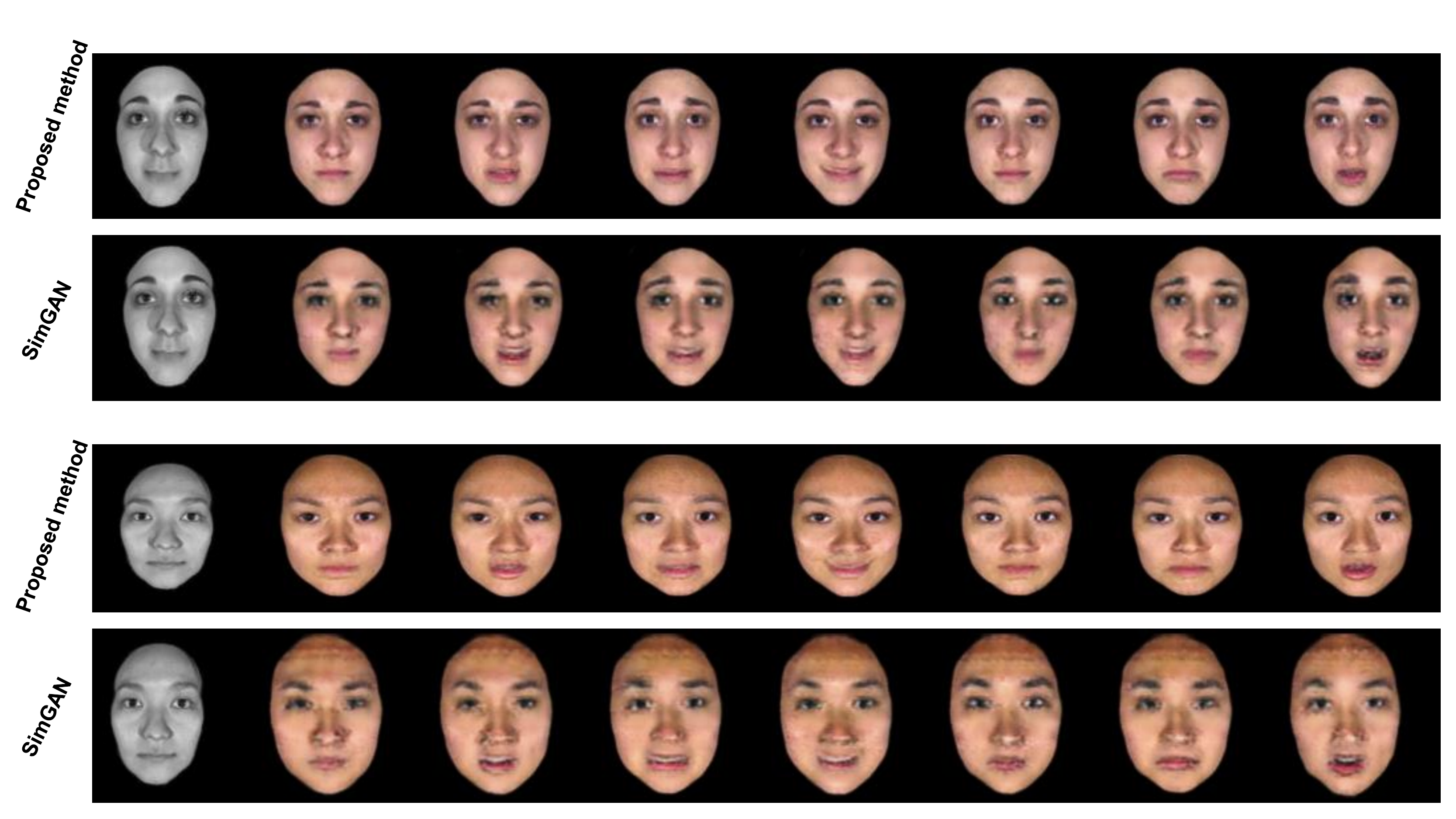}
  \caption{Pose-normalized face attribute transfer results of our proposed method compared with SimGAN method \cite{shrivastava2017learning} on the BU-3DFE dataset \cite{yin20063d}. \textbf{Left to right}: input simulated frontal face and seven different exhibited attributes including \textit{angry}, \textit{disgusted}, \textit{fearful}, \textit{happiness}, \textit{neutral}, \textit{sadness} and \textit{surprised}, respectively.}
\label{fig:6}
\end{figure}

From qualitative results in Fig. \ref{fig:5}, it is obvious that our facial attribute transfer test results (unseen images during the training step) are more visually pleasing compared to other baselines including IcGAN \cite{perarnau2016invertible} and CycleGAN \cite{zhu2017unpaired}. We believe our proposed identity loss helps to preserve the face image details and identity. IcGAN even fails to generate subjects with desired attributes, while our proposed method could learn attribute invariant features applicable to synthesize multiple images with desired attributes.
In addition, to evaluate the proposed realism refinement, the face attribute transfer results of our proposed method have been compared with the SimGAN method \cite{shrivastava2017learning} on the BU-3DFE dataset \cite{yin20063d} (see Fig. \ref{fig:6}).
\subsubsection{User Study}
\begin{figure}[h]
\centering
\includegraphics[height=5cm, width=8.2cm]{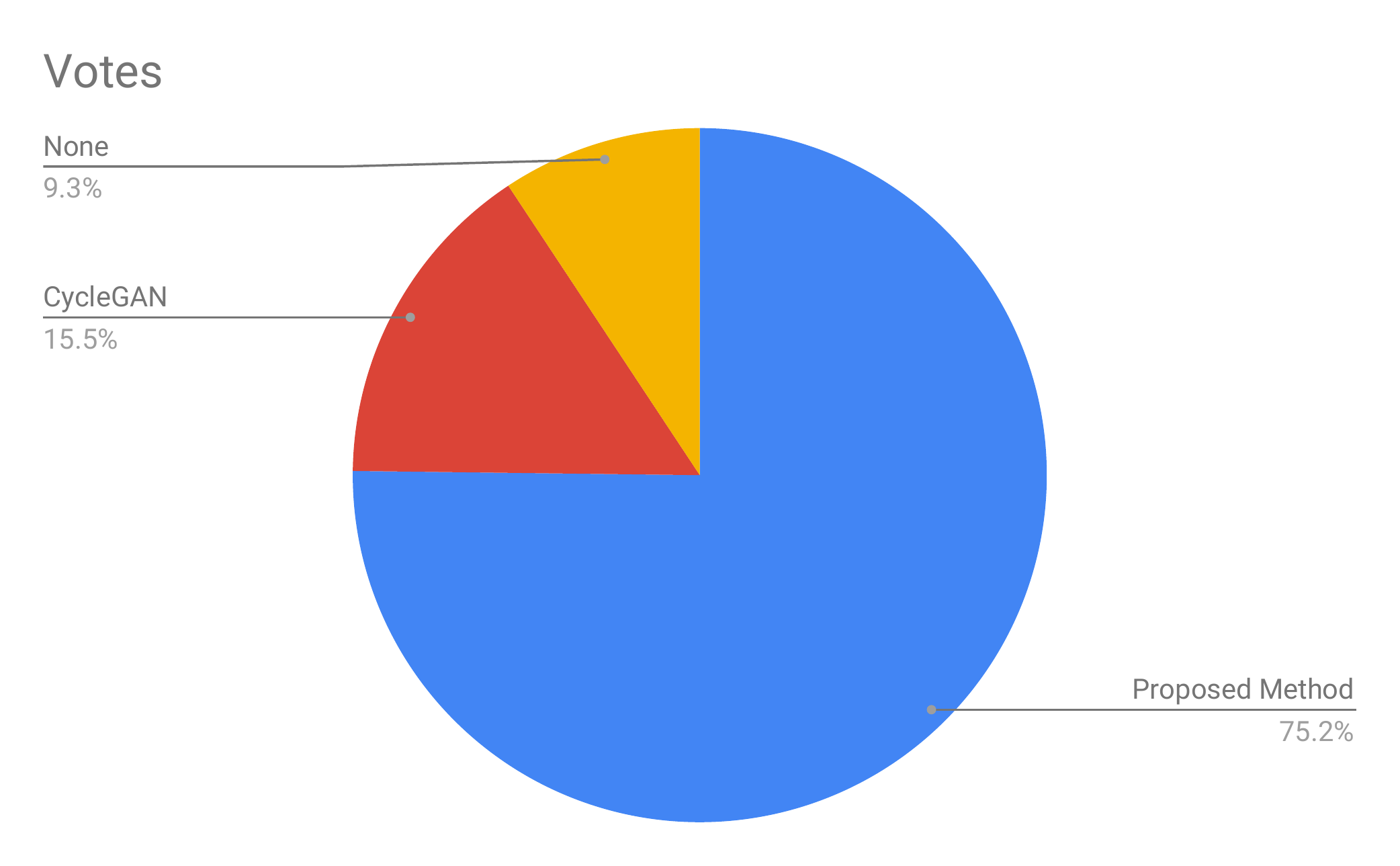}
  \caption{The evaluation results of our user studies to compare our method with CycleGAN \cite{zhu2017unpaired}.}
\label{fig:7}
\end{figure}

We also evaluate the realism of our results with a user study to compare our model with CycleGAN \cite{zhu2017unpaired}. We asked 15 subjects to select results that are more realistic and facial expression is well distinguishable through pairwise comparisons. In addition, a third choice as
``None''  was also introduced in the case if none of them could generate realistic result. 16 random images with the corresponding emotion transfer results from RaFD \cite{langner2010presentation} dataset were presented in a randomized fashion to each person. The Pie chart shown in Fig. \ref{fig:7} illustrates that the results reconstructed by our approach are more appealing to the users.
\subsubsection{Quantitative Evaluation} 
\begin{figure*}[t]
  \begin{minipage}[b]{0.60\linewidth}
    \centering
    \includegraphics[width=0.85\linewidth, height=0.6\linewidth]{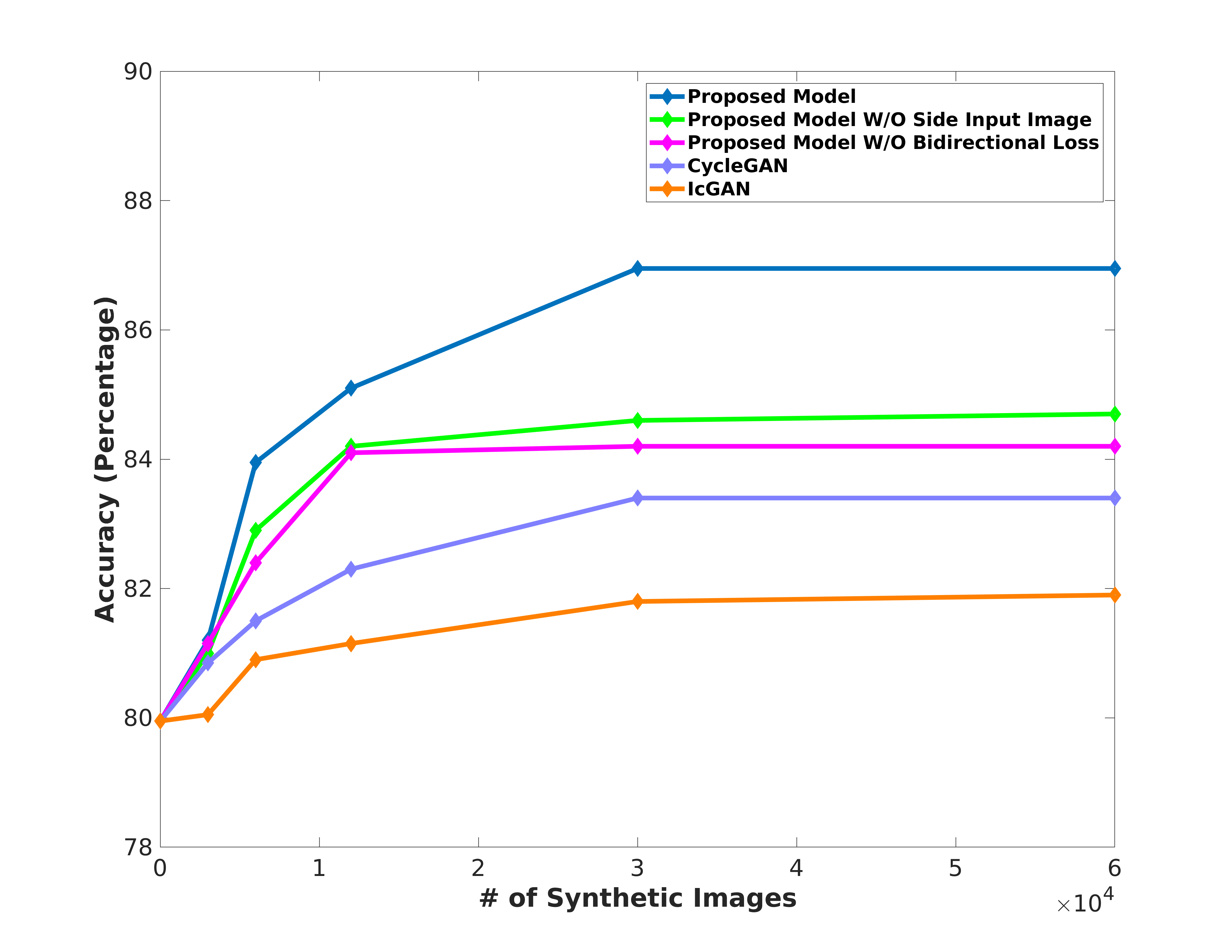}
    \par\vspace{-15pt}
  \end{minipage}%
  \begin{minipage}[b]{0.30\linewidth}
    \centering%
    \begin{tabular}{|c||c|c|}
\hline
Method & Accuracy \\
\hline \hline
HOG 3D \cite{klaser2008spatio}& 70.63\%  \\
AdaLBP \cite{zhao2011facial}& 73.54\% \\
Atlases \cite{guo2012dynamic} & 75.52\% \\
STM-ExpLet \cite{liu2014learning} & 74.59\% \\
DTAGN \cite{jung2015deep} & 81.46\% \\
\hline
\textbf{Proposed Method} & \textbf{86.95\%} \\
\hline
\end{tabular}
    \par\vspace{0pt}
  \end{minipage}
\caption{Impact of the amount of training synthetic images on performance in terms of expression recognition accuracy (\textbf{left}). Performance comparison of expression recognition accuracy between the proposed method and other state-of-the-art methods (\textbf{right}). }
\label{fig:8}
\end{figure*}

We quantitatively demonstrate the usefulness our proposed model in synthesizing photo-realistic facial images controlled by the expression category. Doing so, we augment real images from the Oulu-CASIA VIS dataset with the synthetic expression images generated by our model as well as its variants and then compare with other methods to train an expression classifier. The purpose of this experiment is to introduce more variability and enrich the dataset further to improve the expression recognition performance. In particular, from each of the six expression category, we generate 0.5K, 1K, 2K, 5K and 10K images, respectively. The classifier has an identical network architecture used in synthesizing (RaFD) \cite{langner2010presentation} images except the number of neurons used in the discriminator's fully connected layer. The accuracy results for the expression recognition are shown in Fig. \ref{fig:8}. We can observe that when the number of synthetic images is increased to 30K, the accuracy is improved drastically, reaching to 86.95\%. The performance starts to become saturated when more images (60K) are used. We achieved a higher recognition accuracy value using the images generated from our method than the state-of-the-arts including CNN-based methods e.g., DTAGN \cite{jung2015deep}. This suggests that our model has learned to generate more diverse realistic images. In addition, we evaluate the sensitivity of the results for different components of our proposed method (bidirectional loss and side conditional image, respectively).
Moreover, we evaluate the performance of our proposed method on the MUG facial expression dataset, \cite{aifanti2010mug} using the video frames of the peak expressions. In Table \ref{MUG1}, we report the results of average accuracy of a facial expression on synthesized images. We trained a facial expression classifier with $\left ( 90\%/10\% \right )$ splitting for training and test sets using a ResNet-50 \cite{he2016deep}, resulting in a high accuracy of $90.42\%$. We then trained each of baseline models including CycleGAN and IcGAN using the same training set and performed image-to-image translation on the same test set. Finally, we classified the expression of these generated images using the above-mentioned classifier. As can be seen in Table \ref{MUG1}, our model achieves the highest classification accuracy (close to real image), demonstrating that our model could generate the most realistic expressions among all the methods compared.

For the near IR drivers' dataset, we conducted two set of experiments. In the first experiment, we trained facial expression classifiers with subject-independent subsets (20 subjects for training and validation and 6 subjects for the test). We used multi-view convolutional neural network (MVCNN) \cite{su2015multi} as our baseline. The VGG-Face model \cite{parkhi2015deep} is used as the bottleneck network. In the second experiment setup, we utilized face frontalization scheme and added realisms to the simulated faces using our proposed approach and \cite{shrivastava2017learning}, respectively. As can be seen in Table \ref{psavaleo}, our model achieves the highest classification accuracy, demonstrating that our realism refinement facilitates the synthesized images to preserve much detail of face expression.
\begin{table}[h]
\begin{minipage}[b]{.45\textwidth}
   \centering
   \caption{Performance comparison on the MUG dataset.}
  \begin{tabular}{|c||c|c|}
\hline
Method & Accuracy \\
\hline \hline
Real Test Images & 90.42\% \\
\hline
CycleGAN \cite{zhu2017unpaired}& 84.40\% \\
IcGAN \cite{perarnau2016invertible} & 80.32\% \\
\textbf{Proposed Method} & \textbf{89.91\%} \\
\hline
\end{tabular}
   \label{MUG1}
\end{minipage}\qquad
\begin{minipage}[b]{.45\textwidth}
   \centering
      \caption{Performance comparison on the near IR drivers' dataset.}
   \begin{tabular}{|c||c|c|}
\hline
Method & Accuracy \\
\hline \hline
MVCNN \cite{su2015multi} & 65.35\% \\
SimGAN \cite{shrivastava2017learning} & 69.21\% \\
\textbf{Proposed Method} & \textbf{78.85\%} \\
\hline
\end{tabular}
   \label{psavaleo}
\end{minipage}
\end{table}
Finally, as our last experiment, we performed 5-fold cross validation using 100 subjects for the BU-3DFE dataset \cite{yin20063d}. Training data includes images of 80 (non-frontal face) subjects, while test data includes images of 20 subjects with varying poses. We use the VGG-Face model \cite{parkhi2015deep}, which is pretrained on the (RaFD) \cite{langner2010presentation} and then we further fine-tune it on the frontal face images from BU-3DFE dataset. It can be observed from Table \ref{frontal_cnn} that face frontalization contributes to the expression recognition performance of the profile faces (ranging from 15 to 45 degrees in 15 degrees steps). Having said that, adding realism to the synthetic images (simulated frontal face) helps to bring additional gains in terms of expression recognition accuracy.
 \begin{table}[h]
		\caption{Recognition accuracies on face images at different pose yaw angles.}
		\begin{center}
   \begin{tabular}{|c|c|c|c|c|}
   \hline
\textbf{Method} & \textbf{$\pm 15$} & \textbf{$\pm 30$} & \textbf{$\pm 45$}  \\ \hline
Real Profile Images         & 70.15\%        & 66.50\%        & 58.90\%               \\
Synthetic Frontal Face Images         & 70.91\%        & 65.90\%        & 59.30\%               \\
\textbf{Proposed Method}         & \textbf{72.10\%}        & \textbf{68.52\%}        & \textbf{63.35\%}             \\
 \hline  
\end{tabular}
    \end{center}
    \label{frontal_cnn}
\end{table}

\section{Conclusion and Future Work}
In this paper we propose attribute guided face image synthesis method, which is capable to synthesize photo-realistic face images conditioned on desired attributes. Using our proposed attribute classification objective and incorporating bidirectional learning, we demonstrate a proper way to model latent representation among different domains leading realistic face images as the result. More importantly, we seek to reduce the domain distribution mismatch between synthetic and real faces. In addition, we demonstrate that the synthetic images generated by our method can be used for data augmentation to train face expression classier. We achieve significantly higher average accuracy compared with the state-of-the-art result. In particular, the proposed method surpasses previous approaches by a significant margin of $5.5\%$ on Oulu-CASIA VIS dataset. For the future work, we plan to apply our model to translate dynamic textures of a face from a single image in the context of video domain.





\bibliographystyle{ieee}
\bibliography{egbib}

\end{document}